\documentclass[graybox]{svmult}

\usepackage{type1cm}        % activate if cite above 3 fonts are
\usepackage{makeidx}         % allows index generation
\usepackage{graphicx}        % standard LaTeX graphics tool
\usepackage{multicol}        % used for cite two-column index
\usepackage[bottom]{footmisc}% places footnotes at page bottom

\usepackage{newtxtext}       % 
\usepackage[varvw]{newtxmath}       % selects Times Roman as basic font

\makeindex             % used for cite subject index
\begin{document}

\title*{Beyond transparency: computational reliabilism as an externalist epistemology of algorithms}
% your contribution title if cite original one is too long
\author{Juan M. Dur\'an \orcidID{0000-0001-6482-0399}}
\institute{Juan M. Dur\'an \at Department of Values, Technology and Innovation\\Faculty of Technology, Policy and Management\\Delft University of Technology\\Jaffalaan 5\\2628 BX Delft\\The Netherlands, \email{j.m.duran@tudelft.nl}}
%
% Use cite package "url.sty" to avoid
\maketitle

%\abstract*{This article asks the question, ``what is a reliable algorithm?'' As I intend to answer it, this is a question about epistemic justification. Reliable machine learning gives justification for believing its output. Current approaches to justification (e.g., transparency) involve showing the inner workings of an algorithm (functions, variables, etc.) and how they render outputs. Thus, justification is contingent on what can be shown about the algorithm and, in this sense, internal to the algorithm. In this chapter, I defend an externalist approach to justification, namely, \textit{computational reliabilism} (CR). While I have presented and discussed CR previously in the context of computer simulations (\cite{Duran2013, Duran2018, Duran.Formanek2018}), here I present a more encompassing version, closer to machine learning systems (ML). CR credits reliability to algorithms by identifying \textit{reliability indicators} responsible for the design, coding, use and maintenance of algorithms. Under this heading, we have justification for believing the output of an algorithm when we have identified the appropriate reliability indicators. The main goal of this chapter is to lay the groundwork for CR, how it works, and what we can expect as an externalist epistemology of algorithms.}

\abstract{This chapter is interested in the epistemology of algorithms. As I intend to approach the topic, this is an issue about epistemic justification. Current approaches to justification emphasize the transparency of algorithms, which entails elucidating their internal mechanisms --such as functions and variables-- and demonstrating how (or that) these produce outputs. Thus, the mode of justification through transparency is contingent on what can be shown about the algorithm and, in this sense, is \textit{internal} to the algorithm. In contrast, I advocate for an \textit{externalist} epistemology of algorithms that I term \textit{computational reliabilism} (CR). While I have previously introduced and examined CR in the field of computer simulations (\cite{Duran2013, Duran2018, Duran.Formanek2018}), this chapter extends this reliabilist epistemology to encompass a broader spectrum of algorithms utilized in various scientific disciplines, with a particular emphasis on machine learning applications. At its core, CR posits that an algorithm's output is justified if it is produced by a reliable algorithm. A reliable algorithm is one that has been specified, coded, used, and maintained utilizing \textit{reliability indicators}. These reliability indicators stem from formal methods, algorithmic metrics, expert competencies, cultures of research, and other scientific endeavors. The primary aim of this chapter is to delineate the foundations of CR, explicate its operational mechanisms, and outline its potential as an externalist epistemology of algorithms.}

\section{Introduction} 

%!Phrasing: observations about one or more source domains provide varying degrees of support for a conjecture about a target domain!%

%!increases our confidence instead of justification? increase our confidence to justify our belief in cite results? reasons to believe that cite results are credible? %justified, true and hence assertable belief!%

%% Some algorithms do better than ociters once we restrict cite set of problems we are trying to solve. But no set of norms of deliberation can get us from data to accurate predictions without restricting cite set of problems of interest. I use NFL to bring out non-epistemic aspects of accuracy and ociter measures of goodness of citeories, which helps let go of cite (misleading) intuition that data can be used to choose between citeories in purely epistemic ways. [from https://www.ravitdotan.com/post/my-dissertation-cite-social-view-of-evidence]
% Tiene sentido meter NFL and cite result that non-epistemic aspects of algorithms determine citeir reliability?
%%

The use of algorithms for scientific purposes is delivering remarkable results. A couple of examples will suffice to illustrate this. In molecular biology, AlphaFold can predict protein structures with atomic accuracy in cases where no similar structures are known \cite{Jumper.Evans.ea2021}. In medicine, BenevolentAI has combined structured and unstructured biomedical data sources to identify rheumatoid arthritis drugs like \textit{baricitinib} as citerapeutics for COVID-19 symptoms \cite{Medeiros2021}. In an increasingly number of cases, algorithms have successfully extended cite class of tractable chemistry, biology, physics, and medicine, broadening cite range of modeling and experimental capabilities available to researchers.

Yet, unlike other methods, the algorithm's scientific merits cannot be easily determined by association with a body of scientific knowledge, by adequacy to empirical data, or by diverse theoretical constructs --such as explanation and observation. This is for a variety of reasons. Algorithms are epistemically and methodologically opaque \cite{Humphreys2009}, making it difficult to associate a given algorithm and its output with the general scientific canon. Likewise, empirical phenomena are often temporarily, spatially, or cognitively inaccessible for validation of the algorithm, potentially casting doubts over any representational value of these systems. %; and current efforts to account for cite output of ML face serious shortcomings \cite{Duran2021a}. %cite replication of ML outputs is a citeoretically questionable undertaking, principally because it requires efforts and resources that cannot always be (empirically) secured. As a consequence, citere are significant impediments to making claims about our reliance on citese systems and citeir output.

When confronted with these issues, philosophers and computer scientists gravitate towards \textit{transparency}, an umbrella term capturing diverse methods linking the internal mechanisms and properties of algorithms to their outputs \cite{Creel2020, Wachter.Mittelstadt.ea2018, Ribeiro.Singh.ea2016}. To see how transparency works, consider BenevolentAI. At its core, this algorithm is a search engine that combines structured and unstructured biomedical data sources, drug industry data, and automated retrieval of information from diverse scientific research papers. The data is curated and standardized via data analysis and data fabric. It is then fed into knowledge graphs that structure the data into relationships between diseases, genes, and different drugs \cite{Smith.Oechsle.ea2021}. Richardson led the team that used BenevolentAI to identify rheumatoid arthritis drugs -- notably \textit{baricitinib} -- as suitable therapeutics for COVID-19 symptoms \cite{Richardson.Griffin.ea2020}. To justify Richardson's belief in the scientific value of this output, partisans of transparency would focus on showing how baricitinib is rendered from procedures integrating biomedical data, instantiation of key variables, function calls identifying structural relationships within the algorithm, relevant conditional statements, and other algorithmic operations. Another way to make BenevolentAI transparent is via a \textit{knowledge graph} \cite[30]{Richardson.Griffin.ea2020}. This visualizes how baricitinib inhibits AAK1 (associated with interrupting the COVID-19 virus' passage into cells) and JAK 1/2 (critical for signal transduction pathways), and how baricitinib binds with GAK (known to decrease certain viruses' infectiousness). This knowledge graph also provides reasons to consider drugs like fedratinib, sunitinib, and erlotinib as less effective and, depending on the case, unsafe. For instance, it is shown how these drugs only inhibit AAK1 and neither decrease the chances of cell infection (by binding with GAK) nor inhibit cytokine signaling (by inhibiting JAK 1/2) \cite[30]{Richardson.Griffin.ea2020}.

Are Richardson and his team justified in believing that baricitinib is a medically valid outcome for the issue at hand? What reasons do they have to discard other drugs as either less effective or unsafe? What supports their claim that BenevolentAI is a reliable system for the intended purposes? These are questions about the epistemic reliance on ML and the justification of their outputs. To a great extent, transparency provides answers to these questions. This paper, however, is an effort to provide an alternative answer, one that does not depend on methods for the transparency of the algorithm. More specifically, this paper lays the groundwork for \textit{computational reliabilism} (CR), a reliabilist epistemology centered on algorithms, aimed at justifying their outputs.

As the name suggests, CR borrows bits and pieces from epistemological reliabilism, notably Goldman's process reliabilism \cite{Goldman1979, Goldman2012}. However, the version of CR I develop here draws from, but also expands on, my previous work on computer reliabilism for computer simulations \cite{Duran2018, Duran.Formanek2018}. With these ideas in mind, this chapter is divided as follows. Section \ref{Two-Epi} presents and discusses two epistemologies of algorithms: one that is internal to the algorithm (e.g., transparency) and one that is external to the algorithm (i.e., CR). As expected, these epistemologies have different modes of justification, which are exemplified in section \ref{Example_Wu_Zhang}. The example provided is only intended to motivate CR. In section \ref{CR+}, I lay the groundwork for \textit{computational reliabilism} (CR). Here, I present three types of \textit{reliability indicators} (type-RIs) that credit reliability to algorithms. These are (1) type$_1$-RI technical performance of algorithms (subsection \ref{RI1}), (2) type$_2$-RI computer-based scientific practice (subsection \ref{RI2}), and (3) type$_3$-RI social construction of reliability (subsection \ref{RI3}). In section \ref{final_remarks}, I briefly take stock of my findings and suggest further lines of investigation that substantiate the merits of CR. In gist, this article invites us to reflect on a crucial but often overlooked question: under what conditions are researchers justified in believing algorithms' outputs? My answer is that reliability comes through myriad methods, practices, and processes at diverse stages of specification, coding, use, and maintenance of the algorithms. 

\section{Internalist and externalist epistemologies for algorithms} \label{Two-Epi}

%!I think the new distinction would be ``internal'' where we have the two interpretations below, and ``external'' where we don't know if, without the assistance of the algorithm, we can know the output!%

A central motivation for seeking justification is that algorithms are often epistemically opaque. This concept has two distinct but related interpretations. The first interpretation addresses how algorithms involve multiple complex elements (functions, variables, decisions, data, etc.) in their specification, coding, execution, and maintenance. This means that little can usually be said about how these algorithms cluster data, which criteria are used for creating categories, and overall why algorithms behave the way they do. This interpretation is captured in the epithet `black-box' algorithms as a way to express how far removed algorithms sometimes are from human insight. The second interpretation sets the focus on our limited capacities to say something meaningful about the output of an algorithm \cite[649]{Humphreys2009}. That is, no human being (or group of human beings) can know which functions, variables, decisions, data, etc. are relevant to a given output.

Whereas the first interpretation focuses on the algorithm as an opaque method, the second highlights our cognitive, epistemic, and other limitations in making knowledge claims about the algorithm's output. Both interpretations, I believe, can be cast as human agents lacking proper justification for the algorithm's output. Here justification is taken to be epistemic, and understood in the general sense of a belief being formed in the proper manner. Thus, either because the algorithm is a black-box or because human agents are cognitively limited, there is no basis for claims about the proper formation of a belief about whether the algorithm's output is true, has scientific value, or can be epistemically trusted.\footnote{For simplicity and continuity with the literature on justification, I shall talk of a belief as being ``true'' -- or ``false'' --, scientifically valid -- or sound --, trusted -- or doubtful. However, I will refrain from defending a full-blown realist or anti-realist position. I believe this is a debate that philosophers interested in algorithms need to address at some point -- see the Chapter 3 by Casey in this volume. Thus, to believe that the algorithm's output \=o is to take it that \=o is true, has scientific value, can be epistemically trusted, etc. Following Elgin \cite{Elgin2017}, \textit{truth} will not be understood as (absolute) correspondence with reality, but rather as being sufficiently adequate for the purposes at hand, allowing for practical engagement, scientific progress, and understanding (also \cite{Parker2020a}). Note that talking in these terms doesn't mean that an agent \textit{S} must explicitly believe the proposition that \textit{\=o is true}, since the latter is a different and higher-order belief. That is to say, mere belief that the algorithm's output \textit{\=o} is true doesn't require possession of the concept of ``truth''. Equally important is to note that our beliefs are not necessarily occurrent at any given time, that beliefs come in degrees of strength and confidence, and are historically situated, incremental, and perspectival \cite{Massimi.McCoy2020}. Thanks go to Jack Casey for the close reading of my assumptions and for saving me from making further mistakes.}

Under this heading, transparency surfaces as a promising epistemology of algorithms. It first requires uncovering the inner mechanisms and properties of the algorithm, and then linking these to its output. Human agents are justified by successfully revealing the functions, values, etc., that produced the algorithm's output (\cite{Humphreys2004, Humphreys2009, Alvarado.Humphreys2017, Humphreys2021, Creel2020}). Recall from the introduction that BenevolentAI utilizes \textit{knowledge graphs} to visualize how the algorithm favors baricitinib over other drugs. Having access to how the knowledge graph works and that it rendered baricitinib justifies the belief in the algorithm's output. Another example is LIME: a general algorithm that accounts for the predictions of any classifier by locally learning an interpretable model. Formally, LIME produces a model \textit{g} $\in$ \textit{G}, where \textit{G} is a class of potentially interpretable models (e.g., linear models, decision trees, falling rule lists). In practice, if an algorithm predicts that a patient has the flu, LIME can highlight the symptoms in the patient's history responsible for the prediction. `Sneeze' and `headache', for example, are key variables used by LIME. Indeed, they are flagged as net contributors to the flu prediction. In contrast, `no fatigue' is a variable used as evidence against the prediction \cite{Ribeiro.Singh.ea2016}. Let us note in passing that many forms of explanatory AI (XAI) provide a rich source for transparency, as they often involve tracking back the \textit{path-dependency} of the algorithm that relates a given function (or set of functions), variables, etc., to its output \cite{Duran2021a}.

Thus understood, transparency purports justification as an \textit{internal to the algorithm} matter. That is, the justification of our beliefs that the output is true depends exclusively on some form of surveying the inner workings of the algorithm. To put this idea more or less formally,

\begin{definition}
	A human agent \textit{S} is justified in believing the algorithm's output \=o just in case: a) it is shown, directly or indirectly, the algorithmic path-dependency to \=o; and b) \textit{S} has reasons to believe that the path-dependency to \=o are the case.
\end{definition} \label{Def_T}

Opposing this view is computational reliabilism (CR), here presented as an \textit{external to the algorithm} epistemology consisting of identifying (formal) methods, algorithmic metrics, expert competencies, cultures of research, and the like that make up our best epistemic and normative efforts to specify, code, and maintain reliable algorithms. I shall call these \textit{reliability indicators} (RIs) and, as I shall explain in section \ref{CR+}, they can be divided into \textit{types} and \textit{tokens}. By construction, then, CR does not depend on showing the internal mechanisms and properties of the algorithm. Instead, it depends on reliability indicators that are external to the algorithm. A primer working definition can be,

\begin{definition}
A human agent \textit{S} is justified in believing the algorithm's output \=o if and only if \=o was rendered by a reliable algorithm. A reliable algorithm is one that produces true outputs \=o most of the time. To this end, the algorithm must have been specified, coded, and maintained through diverse reliability indicators.
\end{definition} \label{Def-Reliable-ML}

While I will dedicate a large portion of this chapter to the characterization of type and token reliability indicators, a preliminary conclusion can be drawn now: we can say that transparency and CR have different justificatory modes. According to the former, we have justification by having access to the inner workings of the algorithm. According to the latter, we have justification by identifying methods (formal and otherwise), metrics, expert competencies, cultures of research, and the like external to the algorithm that make up our best epistemic and normative efforts to increase the algorithm's reliability.

As a final attempt to illustrate these two justificatory modes, let me briefly present and discuss an example of an algorithm that classifies individual suspects as \{criminal; non-criminal\} based on their facial traits. It will be shown that transparency justifies the belief that a given suspect is a criminal --or a non-criminal-- whereas CR flags the algorithm as unreliable and therefore lacking justification for such beliefs. It goes without saying that this example is only meant to contrast these two justificatory modes. No conclusions about their individual value as epistemologies are intended to be derived from it.

\subsection{Merchants of mistrust} \label{Example_Wu_Zhang}

%% An example where transparency fails %%

In 2016, computer scientists Xiaolin Wu and Xi Zhang developed a Convolutional Neural Network (CNN) that analyzed over 1,850 ID photos and classified them as \{criminal; non-criminal\}.\footnote{It is worth noticing that Wu and Zhang’s use of photos from actual people, in contrasts with other approaches that use synthetically generated photos \cite{Turk.Pentland1991, Blanz.Vetter1999}.} About 1,120 of these photos were of people with no criminal convictions, and the remaining were of people who were either wanted for crimes or convicted of crimes. The CNN's operation was simple. It picks out facial traits (e.g., distance between the eyes, length and curvature of the mouth) and classifies each photo as \{criminal\} or \{non-criminal.\}. No other concept or category was operational. Despite this -- or perhaps because of this -- the predictive accuracy measured using the Area Under the Receiver Operator Characteristic Curve (AUC-ROC) was very impressive: Wu and Zhang measured 0.9540 accuracy in the classifications. This means that the CNN was able to successfully classify faces of individuals as being \{criminal\} or \{non-criminal\} approximately 95\% of the time \cite[2]{Wu.Zhang2016}. 

To further validate their algorithm and rule out that such a high predictive accuracy resulted from overfitting, Wu and Zhang retrained the CNN on a dataset where the labels ‘criminal’ and ‘non-criminal’ were assigned randomly as negative and positive instances with equal probability. For the retraining case, the CNN failed to distinguish between the two categories, plummeting the average classification's accuracy to 48\%, with a false negative rate of about 51\%, and the false positive rate close to 50\%. Wu and Zhang also accounted for problems related to unbalanced datasets, choice of photos (light, angle, over and under exposure, clothing, etc.), and other issues pertaining to accuracy. To most algorithmic standards, these results speak in favor of a reliable CNN capable of consistently classifying the photos in question.

Wu and Zhang naturally defend the scientific merits of their algorithm. To their mind, as to many, high predictive accuracy means that the algorithm's outputs have scientific value, are true, etc. It is thus no coincidence that they confidently announce the “law of normality for faces of non-criminals” \cite[8]{Wu.Zhang2016}. But high predictive accuracy is no standard for claims about scientific value or truth. One could argue that while the Ptolemaic model exhibited high predictive accuracy in its measurements, the model fundamentally misconceived and misconstrued planetary motion. Additionally, we know that predictive accuracy can be manufactured by carefully selecting the input data and calibrating variables and functions in the algorithm to some desired degree. For example, finding optimal values for hyperparameters (number of hidden layers, batch size, choice of activation function, etc.) is fundamental for having faster convergence, high accuracy, and overall better results. Now, algorithms allow multiple optimal hyperparameter configurations depending on datasets, purposes of the algorithm, and tasks \cite{MoralesHernandez.VanNieuwenhuyse.ea2022}. Furthermore, optimal configurations for one algorithm do not typically translate to others, making them incompatible in many different ways \cite{Rijn.Hutter2018}. As a result, selecting optimal values for hyperparameters, along with the best configuration for a given algorithm, is largely a matter of human decision. Without further provisions in place, such as ensuring compliance with scientific standards, professional integrity, and standardized measurements for the optimality of hyperparameters, predictive accuracy can (relatively easily) be manufactured. 

These authors would insist that high predictive accuracy grants scientific value to their CNN's outputs. To further defend this, they retrained the parameters of every layer in the CNN while also modifying the architecture \cite[3]{Wu.Zhang2017}. As a result, the high accuracy in the output remained at the same levels. In fact, the CNN correctly picks out specific facial attributes from photos, and then classifies them into the appropriate category ca. 95\% of the time.\footnote{Despite these efforts, the system's high accuracy remains questionable. While there is no evidence of output manipulation, one can't help but wonder whether the system would maintain the same level of accuracy when faced with a larger and more diverse datasets.} But again, taking high predictive accuracy as an indication of the reliability of automated inference on criminality algorithms is problematic. It confounds justification of a technically correct output with the justification required for believing that output.\footnote{What is operating here is the distinction between output accuracy, which is concerned with the correctness of the final results produced, and procedural accuracy, which is concerned with the execution of steps and adherence to methods.} Wu and Zhang have no justification for believing that someone is a criminal based on facial traits alone. This is the case regardless of how accurate their algorithm is at picking out and classifying photos. 

In this context, transparency does not seem to be of much help for justification. When Wu and Zhang try to justify their outputs on high predictive accuracy, they look at what their AUC-ROC values are telling them. This means that specific inner functions and properties of the CNN responsible for the output will support the justification. But justifying the CNN's output using the same functions used to produce them is epistemically circular and inadmissible for the proper formation of beliefs. Rather, these functions only speaks of the algorithm's robutness, and only in a very limited way. It follows that Wu and Zhang can pin down the functions and properties of their CNN that account for the high predictive accuracy, but at no point can they use those functions and properties alone for claims about justification. In other words, transparency here does not help to distinguish what we are compelled to believe from what cements that belief.\footnote{As suggested earlier, transparency is a broad concept that admits different interpretations. A partisan of \textit{post-hoc} explanatory AI, for instance, could argue that the algorithm was not taking into account scientifically salient aspects of the pictures. By means of this, one could in principle identify high-level features that refer to domain knowledge (e.g., a list of criminality-based characteristics) and thus have grounds -- reasons, evidence -- for claims about justification.}

\section{Computational reliabilism (CR)} \label{CR+} 
%As will become clear when discussing RIs, my thesis supports the idea that belief, justification, and knowledge 

%TO BE JUSTIFIED UNDER CR: It is our current best option because, first, the reliability indicators draw from making our best epistemic efforts to justify our belief in \=O. To be justified here means that \=O upholds to standards of scientific knowledge and practice, and that it can be reasonably taken as the basis for current scientific inquiry and possibly as a landmark for further research. It also means that \=O is scientifically objective in the sense that it relates the justification of my beliefs in \=O with a strong sense of endorsing this justification to others \cite[116]{Douglas2009}. Second, because CR can identify reliance and the ``extra factor'' necessary for trustworthy ML.s

Claims about justification find a home in CR, a branch of process reliabilism where subject \textit{S} is justified in believing output \=o if the algorithm is reliable (see definition \ref{Def-Reliable-ML} on page \pageref{Def-Reliable-ML}) \cite{Duran2018, Duran.Formanek2018, Goldman2012}. An algorithm is reliable when it produces \=o that are true rather than false most of the time. It is important to note that reliability here is not merely a matter of track record but rather about the algorithm's propensity to generate true =o in most cases. Now, the debate in epistemology over the most suitable version of reliabilism is extensive and cannot be addressed here. Suffice it to say that I favor \textit{propensity reliabilism} over Goldman's \textit{frequentist reliabilism}, aligning with Alston, for whom ``[a] reliable instrument is one that \textit{would} usually deliver favorable results over an appropriate range of cases \textit{if and when} they occur'' \cite[6]{Alston1995}. I will not discuss this point further as the relative frequency or propensity of CR are unproblematic for the purposes of this chapter. The question is rather, how to confer reliability to an algorithm. To achieve this end, CR utilizes \textit{reliability indicators} (RIs) as markers of methodological, cognitive, social, and epistemological competence. RIs are any algorithmic-related methods, metrics, practices, domain-specific knowledge, and the like with a reliability-conferring property. Although I will neither discuss the nature of this reliability-conferring property nor how it operates, a simple example should illustrate that it is not of a `spooky' kind --rather, it is very familiar to many philosophers of science \cite{Kitcher1993}.\footnote{This is, of course, not to say that CR faces problems (See Chapter 5 by Alvarado in this volume).} Consider the microscope. Claims about its reliability stem from, say, the use of the laws of optics for its construction and calibration, the effective observation of entities also dependent on the researcher's prior knowledge, and a scientific community with the background education and capacity to accept or reject an observation. The proper workings of the instrument, the knowledge of the right methods, and the social validation of an observation all confer reliability to the microscope. The RIs I shall discuss shortly play similar reliability-conferring roles, as they amount to accessible scientific practices, methods, cultures of research, scientific debates, and other (more or less) scientifically-grounded activities. Nothing spooky about that. The real challenge, rather, is to be as precise as possible in identifying RIs for specific cases. Here is where this chapter falls short. However, this is for a good reason. Recall that my only pretense with this chapter is to lay down the groundwork for an externalist epistemology of algorithms, and therefore my treatment of CR will be very general. The reader interested in concrete applications of CR to different domains is cordially invited to read \cite{Duran.Jongsma2021} for cases on medicine and healthcare, and \cite{Duran.Vloed.eaunderreview} for forensic science. 

\subsection{Reliability Indicators}

For conceptual clarity, I distinguish between \textit{type}-RIs and \textit{token}-RIs. While the former refers to a unique category of indicators, the latter refers to an individual occurrence for that category. With this distinction in mind, the following type- and token-RIs are at the heart of CR:

\begin{itemize}
	\item \textit{Type$_1$-RI - Technical performance of algorithms} focuses on the specification, coding, execution, maintenance, and other technical features that contribute to the performance of the algorithm (e.g., high accuracy and low rate of errors, but also tolerance to domain change, repurposability, reusability, modularity, etc). In this sense, typical cases of token$_1$-RI include practices and protocols for collecting, curating, storing, distributing, and analyzing data; the use of out-of-distribution data and data augmentation, parametrizations; bechnmarking; choice of architecture; treatment of algorithmic kludges \cite{Clark1987}; recasting \cite{Duran2020}; error treatment, and other techniques pertaining to achieving the desired performance of algorithms. Within this type-RI could also be included a justification for the employment of said practices, metrics, and methodologies, along with the specific circumstances in which the algorithm is specified and coded.
	\item \textit{Type$_2$-RI - Computer-based scientific practice} focuses on securing algorithmic-based scientific research. It results from the operationalization and implementation of scientific concepts, causal structures, models and theories, laws and law-like principles, taxonomies, but also scientific metaphors and intuitions, values (epistemic and otherwise), idealizations, abstractions, and representations. This type-RI intends to capture the degree to which scientific units of analysis are implemented and operationalized into the algorithm. The selection and justification of domain knowledge are equally crucial in enhancing an algorithm's reliability. Let us note that the viability and success in doing so largely depend on algorithmic-related decisions, such as programming language choice, the use of formal techniques like verification methods \cite{Fetzer1998}, and the utilization of sub-modeling and multi-modeling \cite{Duran2020}.
	\item \textit{Type$_3$-RI - Social construction of reliability} focuses on broader goals related to accepting -- or rejecting -- algorithms and their outputs by diverse communities (e.g., scientific, academic, the general public), the realization of intended values and goals, and the overall assessment of the algorithm's scientific merits. This occurs through token$_3$-RI such as debates, experimenting and testing, replicability of results, and other forms of intellectual exchange.\footnote{Let me echo what Heather Douglas \cite{Douglas2004} persuasively argued: scientific and computational practices, as presented in type$_1$- and type$_2$-RI, along with the social processes tailored to them, as presented in type$_3$-RI, are neither reducible to one another nor completely uncoupled. This sentiment applies here as well.}
\end{itemize}

Under this heading, CR is understood as a family of reliability-eliciting algorithmic-related indicators capable of crediting an algorithm as a reliable belief-forming method. It is important to note that by accepting a reliabilist epistemology, one also accepts the propensity likelihood that governs the reliability of a process. This translates into acknowledging that algorithms can occasionally be inefficient, contain errors, be unsuitable for specific purposes, misrepresent, and compute incorrect results. If failures perpetuate over time, the relative propensity governing CR shifts, rendering the algorithm ultimately unreliable.

Furthermore, proponents of CR take note of human cognitive limitations in accessing some token-RIs, which conditions the claims about the reliability of an algorithm. They also need to accommodate the fact that token-RIs are neither absolute nor universally applicable. Not all token-RIs are credited, relevant, and applicable under the same criteria, nor does the same token-RI equally apply to all algorithms. CR is thus understood as perspectival, provisional, and subject to corrections, with no particular token-RI considered to have an all-or-nothing reliability-conferring property.

Thus understood, token-RIs come in degrees. The degree to which one token-RI is more relevant than another, or contributes to the overall reliability of the algorithm will depend on the context in which the algorithm is specified, coded, used, and maintained. It will depend on the epistemic and non-epistemic values and goals at stake. It will also depend on the culture of specifying, coding, maintaining, and using the algorithm of a given community \cite{Sundberg2010}. In this sense, no individual (set of) token-RI can guarantee the reliability of all algorithms. Furthermore, even under the assumption that some token-RI is suitable for a given algorithm, this does not ensure that our reliability claims are eternally warranted. Old token-RIs can lose their appeal as new ones come to light. For these reasons, this chapter holds no pretensions to claim the completeness of the various type- and token-RIs presented here. Further arguments could be given on the need for additional type- or token-RIs not discussed here, or that some indicators are somewhat misplaced, or that some others need replacement. None of this is to say, however, that there are no stable type- and token-RIs that apply across many reliable algorithms. In fact, most of the token-RIs discussed next maintain, to my mind, a permanence in time despite changes and fine-tuning that occur with new technological and scientific developments.

Finally, I recognize that CR may not be readily accepted by everyone. As a reliabilist epistemology, one might feel that it still needs to address a few concerns. For starters, there are issues pertaining to the relevance and availability of type- and token-RIs, potential conflicts emerging among token-RIs, and their precedence, order, and weight. Unfortunately, these issues will not find a complete answer here. To my mind, it is the richness and urgency of this problem that requires putting into practice demands for an account at least as complex as the one presented here. In this sense, CR does not provide, nor intend to provide, absolute assurances. Instead, CR aims to highlight that our best epistemic efforts can be geared towards the reliability of algorithms. Little more can be expected given the fallibility and limitations of human cognition. Taking note of these caveats, I now discuss a few types- and token-RIs in more details.

\subsubsection{Type$_1$-RI: Technical performance of algorithms} \label{RI1}

In earlier versions of CR \cite{Duran2013, Duran2018, Duran.Formanek2018}, RIs mainly focus on the specification, implementation, tractability, and overall performance of algorithms. For instance, the first three token$_1$-RI discussed in \cite{Duran.Formanek2018} put forward defining criteria for assessing the utility value of algorithms and their outputs \textit{qua} computational methods. Consider \textit{validation} procedures as an example.\footnote{It is important to recognize that various forms of verification and validation exist, each conferring different degrees of reliability. Decisions must be made \cite{Oberkampf.Roy2010}.}. In automated diagnosis, algorithms are used for patient prognosis. One way to increase our confidence in the output is to compare the disease progression as indicated by the algorithm with clinical data from prior patients that share the same endotype or phenotype \cite{Myszczynska.Ojamies.ea2020}. This practice validates the synthetic data rendered by the algorithm with empirical data collected via diverse scientific methods (e.g., observation, experimentation, intervention, measurement, and others). The utility value of the algorithm is then considered appropriate if validation standards are satisfactory. 

In this respect, subjecting algorithms' outputs to validation methods increases -- or reduces -- our confidence in the reliability of an algorithm, as it is a good indication of the algorithm's accuracy and margin of error. Validation methods also give a fair sense of the capacity to generalize the algorithm from the training data to new, undiscovered data. From a scientific perspective, validating algorithms also contributes to the rigor and reproducibility of research, ensuring that findings are based on sound methods.

Now, it should be expected that validation methods encompass a variety of techniques and methods.\footnote{See Chapter 14 by Manganini and Primiero  in this volume.} As such, they are not all appropriate for the same goals. This means that a given validation techniques cannot be simply applied to different algorithms without prior critical discussion. There must be agreement on how suitable a given validation technique is for the algorithm and data in question, as well as the purposed goals and tasks \cite{Lorscheid.Heine.ea2012, Fagiolo.Moneta.ea2007}. This is an often overlooked aspect of the social dimension of engineering the performance of algorithms. In \cite{Duran.Formanek2018}, we argued for a \textit{history of (un)successful implementations} that affords this interpretation. The idea is simple and intuitive: good practices with visible success --such as high accuracy, low margin of error, ease of implementation, and formal verification-- tend to endure over time, while less successful practices tend to be eradicated.

To illustrate this token$_1$-RI a bit further, consider \textit{design prototyping}, a sub‐field of software engineering that assists developers in assessing alternative design strategies and deciding which is best for a particular goal. Since there are no standard methods for choosing the best strategy, researchers need to compare the requirements of the algorithm with various design approaches to evaluate which one possesses the best characteristics for fulfilling the intended objectives. The example I used in previous publications is a computer simulation involving networking. For this, there are different topologies: ring, star, tree, and mesh. In order to pick the most suitable one, diverse performance characteristics need to be evaluated to see which topology is better at meeting performance goals and constraints \cite[Chapter 5]{Pfleeger.Atlee2009}. 

The same point can be made with an example closer to machine learning. Take the case of BenevolentAI presented earlier, which utilizes \textit{Best First Search} (BFS), a search algorithm highly successful for navigating graphs and trees. The primary goal of BFS is to find the most promising path to a target node based on a given heuristic. In this respect, BFS has proven to be extremely effective for searching suitable drugs within BenevolentAI's knowledge graph \cite[604]{Segler.Preuss.ea2018}. Classified under type$_1$-RI \textit{history of (un)successful implementations}, BFS contributes to the reliability of BenevolentAI and the justification of its outputs.

Likewise, past failures must be, and typically are, avoided by competent programmers. The history of computing is littered with cases of failed software that changed specification and coding practices. Therac-25 is one tragic case \cite{Leveson.Turner1993}. As reported, the algorithm used by Therac-25 was not thoroughly validated, and the testing process was insufficient to catch critical bugs that led to radiation overdoses. Furthermore, there was poor error handling and reporting in the software. Error messages were often cryptic, and operators were not adequately trained to understand and respond to them. Finally, there were no redundancy safety mechanisms that could ensure that software failures do not result in such catastrophic outcomes. From the perspective of CR, these all amount to diverse indicators of the unreliability of the algorithm used in Therac-25.

\subsubsection{Type$_2$-RI: Computer-based scientific practice} \label{RI2}

Assessing the technical performance of algorithms facilitates justification in terms of increasing accuracy, predictive power, low error rates, tolerance to domain change, and the ability to multi-purpose algorithms and data, among other factors. However, this assessment is silent on the adequacy of algorithms for scientific purposes. A reliabilist epistemology must offer standards by which the algorithm used in a scientific context can be warranted to a greater or lesser degree.

Let me illustrate these ideas with a familiar example. Wu and Zhang's automatic facial recognition system exemplifies how accuracy alone does not exhaust the reliability of an algorithm. As mentioned in section \ref{Example_Wu_Zhang}, the AUC-ROC measured 0.9540 predictive accuracy for their CNN. Such tremendous results cemented these researchers' confidence in the scientific merits of the algorithm. However, as discussed, there is no basis for such optimism. Criminality is a socially constructed concept that depends on diverse and sometimes contradictory interpretations of the socio-economic basis of criminality, psychological studies of criminals, and laws that determine when and to what degree someone is considered a criminal. Without reference to some of these concepts and frameworks, the prediction --however accurate-- lacks the grounds for legitimate scientific claims.

Under CR, the reliability of algorithms is not exclusively assessed based on high predictive accuracy. Science involves more than just measuring and classifying algorithmic outputs. In this respect, I believe that algorithms cannot and should not operate in isolation from the broader context of scientific undertakings. We need to delve not only into standard non-algorithmic scientific practice, but also into a form of scientific practice that evolves with and heavily depends on algorithms. Type$_2$-RI is an attempt to capture the family of token-RI connected to a larger body of scientific theories, beliefs, and practices within which algorithms are specified, coded, utilized, and maintained. In what follows, I lay out two potential candidates.

\paragraph*{Expert knowledge} \label{Expert_Knowledge}

\textit{Expert knowledge} is an umbrella term that covers the myriad of background education, knowledge, activities, training, virtues, and skills of researchers that bring to bear a broad range of talents to the specification, coding, use, and maintenance of algorithms in scientific contexts. Understood as a reliability indicator, \textit{expert knowledge} reports on the many ways in which scientific expertise, technical expertise, and general competencies can be implemented into an algorithm.

To best understand this indicator, we must look at its various functions. For starters, it puts forward the algorithm's competencies, scope, and theoretical assumptions as conceptualized by the researchers involved in the specification, coding, maintenance, and execution of the algorithm. It also accounts for the ability to describe a target system and its conditions for adequacy (e.g., to be applicable in a specific domain, to be representative of a particular condition, to be context-sensitive, to be repurposed). Expert knowledge covers social practices tailored to the development of algorithms, aptitudes to anticipate their merits intelligibly, and abilities of agents to manipulate them. For instance, setting up the variety of initial conditions, datasets, parameters and hyper-parameters (epochs, batch size, number of neurons, number of layers, dropout rate, etc.), all of which are complex yet critical for the performance and scientific merits of the algorithm. Consider determining which parameter to prioritize as a reliability indicator. Their selection and optimization are not trivial and yet fundamental for the general performance of algorithms (convergence of results, accuracy, overall performance) \cite{Hutter.Hoos.ea2014}. van Rijn and Hutter have conducted an informative experiment to show that the final performance metrics for deep learning models vary according to how different researchers select and optimize algorithmic parameters and instantiations \cite{Rijn.Hutter2018}. Thus understood, experts contribute to the overall reliability of an algorithm by specifying relevant internal data-types, structures, relations, operations, and the like. They also credit reliability (or might identify instances of unreliability) by their pick and choose of datasets, parameters, and other variables. 

As a reliability indicator, expert knowledge also attempts to accommodate the complexities of algorithms through the division of cognitive labor. Rather than being developed in isolation, algorithms involve a myriad of direct and indirect stakeholders (e.g., software engineers, physicians and chemists -- in the case of BenevolentAI --, biologists -- in the case of AlphaFold --, and psychologists and legal officers -- in what should have been the case for Wu and Zhang). A core team specifies and codes algorithms utilizing ready-made computer modules others have coded. They employ measuring techniques others have designed, constructed, and calibrated. They analyze data using mathematical and statistical techniques others have validated. They make use of mathematical and computational methods others have devised and tested. There is no development of algorithms in solitude. Teams with diverse cognitive strengths and talents collectively collaborate in a variety of ways. Hence, the success or failure of algorithms is tailored to this collective knowledge, just as much as it depends on individual competencies. What one team member overlooks, another might notice. What one team member forgets, another might foresee. What one team member does not know how to solve, another might be able to teach. Thus diversified, the range of achievable solutions is far greater than what is available in atomized practices.

Interestingly, the role and value of experts are being increasingly recognized in philosophical studies on algorithms. Ratti and Graves \cite{Ratti.Graves2022} argued that documenting developers' motives and the code and specification of an ML are indicators of the reliability of the system. Newman \cite{Newman2016} has argued along similar lines with respect to computer simulations. Newman considers the entire practice of software engineering to be at stake, from test plans to selecting programming languages and modeling tools, including configuration management.

While I am sympathetic to these ideas, my interpretation of expert knowledge is somewhat broader. It includes technical personnel with no training in software development, practices that exceed software engineering standards, and accommodates the possibility that complete documentation of an algorithm is not always available.

In practice, non-technical personnel are intimately involved in algorithmic development (e.g., physicians and chemists in the case of BenevolentAI, and biologists and chemists in the case of AlphaFold), despite having little to no idea how key features of the system are specified and implemented. Their expertise is, however, crucial for the assessment of the reliability of the algorithm, and thus must be considered. Typically, their role is to inform, supervise, and sometimes even test the specification and coding of algorithms. But of course, these roles and interactions vary among cultures of research \cite{Sundberg2010}.

In connection with this, local practices and vernacular terminology often exceed what is captured by software engineering standards. Consider for instance how ML naturalizes or ‘fossilizes’ concepts. Once a concept is coded into the system, it is universally and indistinguishably applied across large and heterogeneous databases with varying degrees of success. Take the concept of ‘health’ as a case in point. One interpretation takes statistical measures and standards of normal biological measurements of someone's body as the baseline for whether they are healthy. This concept of ‘health’ can be relatively straightforwardly implemented on an algorithm. However, the same concept also allows interpretations tailored to the diverse values of an individual or a community \cite{Richman2004}. If a community considers blood transfusion to be harmful, they will treat any members of the community who have received a blood transfusion as unhealthy \cite{Richman.Budson2000}. Implementing a cogent definition of health is no trivial matter.

Lastly, anyone who has written a piece of code knows all too well that not every line of code is documented. And even if algorithms were fully documented, this is no guarantee of understanding the code and its various functions. Thorough documentation -- when it happens -- and well-intended software engineering might still fall short of capturing the methodological and epistemological competencies of algorithms. We need to highlight the subtle interpretations, gentle disagreements, and non-verbal practices pervading computational and scientific practice and which make their way into the algorithm.

Let me finish by noticing that this reliability indicator brings about another important aspect of algorithms, namely, that they might only be \textit{locally} reliable. The idiosyncrasies attached to documenting, specifying, coding, executing, and maintaining algorithms might make them only reliable in one context but not necessarily in another. This is, I believe, at the root of IBM's Watson for Oncology's difficulties of implementation in South Korea and Denmark, despite its success in the US market \cite{Vulsteke.Arevalo.ea2018, Emani.Rui.ea2022}. Notoriously, Watson for Oncology was capable of analyzing large amounts of data and multiple variables, rendering accurate diagnoses and treatments for cancer patients in the US. But while IBM presents Watson for Oncology as offering more objective medical decisions and more accurate diagnoses than actual oncologists \cite{Swetlitz2016}, it has been reported that many of these claims have been aggrandized \cite{Ross.Swetlitz2017, Ross.Swetlitz2018}. When implemented in South Korea and Denmark, only a fraction of the outputs rendered by the algorithm matched -- or closely matched -- the local clinician’s best diagnosis \cite{Hamilton.GenoffGarzon.ea2019}.

%To reiterate the same idea utilizing Wu and Zhang's ML. It is clear that this system misconstrues the notion of criminals and criminality with potential dire implications: it will lead to unjust sentencing, damage self-perception \cite{Deery2017}, and ultimately undermining the rule of law, weaken the social fabric, and erode citizens' trust in institutions. Whereas the practice of documenting and software engineering are suitable for detecting and backtracking failures and unintended behavior in the system, by themselves they are inadequate for justifying our belief in the output of ML.\footnote{It is possible to relate these practices with Creel's interpretation of transparency \cite{Creel2020}. According to the three levels of transparency discussed by the author, at least two can be approached by software engineering practices, namely, algorithmic transparency and structural transparency.} 

\paragraph*{Knowledge-based integration} %Embedding (within) scientific knowledge}

%!So, to ascertain what, say, the second law of thermodynamics amounts to, we must see how it enters into explanations, how it bears on experimental design, how it is integrated into theory, and so on. We must see, that is, how unquestioning allegiance to it affects the ongoing activity of a scientific community. (Elgin, 75)!%

%The good reason to believe a scientific claim is a matter of overall fit into a coherent network of other claims, theoretical and empirical. This means that no one can responsibly judge the plausibility of a scientific claim quickly or in isolation from its broad context. There is no short course to scientific expertise, no short cut to making informed decisions. I have to admit that I am in no position to judge for myself whether string theory is likely to be true. I have to leave this up to those who have spent the time learning the whole story. For the same reason, school children are in no position to judge for themselves whether the theory of evolution is likely to be true. Scientific judgment is informed judgment, and that requires knowing a lot of the context.

Scientific results do not come in discrete bits, nor are the objects of scientific inquiry independently sanctioned. Instead, scientific theory and practice constitute a web of mutually supportive claims and commitments that are reached after complex negotiations in complex socio-economic and political environments. However, many studies utilizing algorithms portray a sanitized image of scientific research, where there is privileged access to structured data, undisputed model implementation, and meaningful representations of the world.  %This, with supporting evidence that the system implements theories on criminality, criminal psychology, and social studies on crime. The expertise of Wu and Zhang is funneled into the ML as concerns to finding the most functions that pick out facial traits in large -- presumably unpolluted -- datasets. It is, in fact, 

Wu and Zhang, for example, state that the quality of their databases and the methods implemented for data analysis prevent ``the garbage of human biases from creeping in'' \cite[2]{Wu.Zhang2017}. Given ``race, gender and age, the faces of [the] general law-abiding public have a greater degree of resemblance compared with the faces of criminals'' \cite[2]{Wu.Zhang2017}. It is, however, doubtful whether Wu and Zhang's CNN has any scientific merits. One reason (to add to those previously mentioned) is that Wu and Zhang's CNN is largely disconnected from accepted bodies of scientific knowledge. More precisely, the categories their CNN purports to use (i.e., \{criminal\} and \{non-criminal\}) are posited in isolation from established evidence, models of criminal psychology, social studies on crime, and the relevant theories on criminality. Thus, the CNN's outputs are based solely on picking out facial traits from selected photos, rather than being premised on a larger body of knowledge implemented in the algorithm. %If anything, Wu and Zhang's CNN supports discredited theories such as phrenology, genetic reductionism, and biological determinism.\footnote{To put the same idea in counterfactual form: had Wu and Zhang framed their CNN into a larger body of scientific knowledge, it would have provided unsupported evidence on criminality.} This latter realization clearly does not provide justification for believing in the CNN's output.

%Maybe it is not that the model doesn’t have a theory to back it up, but that the jbda reaches beyond what the data says.

I will call approaches that conceive of algorithms as disconnected from the larger body of scientific knowledge \textit{just a bunch of data analysis} (JBDA). By doing this, I intend to emphasize that an algorithm performing mere data analysis, but disconnected from concepts, theories, law-like principles, hypotheses, and other scientific units of analysis, is unlikely to merit scientific credentials, regardless of its predictive accuracy. To my mind, JBDA ignores the `bigger picture' of knowledge integration, interpretation, and operationalization into algorithms. In fact, I consider JBDA as misleadingly portraying algorithms as an objective, unambiguous, and scientifically grounded examination of data that produces scientifically meaningful outputs. Nothing could be further from the truth. Wu and Zhang's CNN approach is an archetypal JBDA, as it depicts scientific practice as granular, consisting of discrete pieces of information, separately secured and individually sanctioned. To these authors' minds, ``like most technologies, machine learning is neutral'' \cite[2]{Wu.Zhang2017}. JBDA approaches advocate a form of scientific practice that is non-perspectival, socially disinterested, and impartial (i.e., epistemically and normatively neutral\footnote{For a critical view on this perspective in the context of algorithms, see \cite{Pozzi.Duran2024}.}), and disembodied from a larger corpus of scientific knowledge.

Are there instances where JBDA approaches are scientifically intelligible? I believe so. As suggested, there are indeed cases where mere data analysis renders valuable scientific insight about a subject matter. But for such cases, one needs to provide further justification that relates JBDA with a larger corpus of knowledge. A plausible interpretation of an account of scientific practice with algorithms capable of accommodating cases of JBDA takes the bulk of scientific knowledge and practices in the field under study as background knowledge and as affording sufficient grounds to underwrite particular claims made with the algorithm.\footnote{Helen Meskhidze makes a similar claim using ML applications in astrophysics. The chapter uses ``physics-informed machine learning'' where physical laws and domain-specific knowledge implemented in the algorithm are crucial for its success. In terms of an epistemology of algorithms, my approach differs in that Meskhidze is interested in fostering transparency and interpretability (See Chapter 18 by Meskhidze in this volume).} To briefly illustrate this idea, as this point encroaches on issues discussed under type$_3$-RI (see section \ref{RI3}), consider BenevolentAI*, an ML system whose working principles are JBDA. Suppose that BenevolentAI* puts forward baricitinib* as a drug with high chances of combating COVID-19 symptoms. Would researchers be justified in believing baricitinib*? Surely not at face value, but only once it is embedded in a larger body of knowledge about COVID-19 and after some clinical trials show its scientific worth. %This interpretation, however, shifts the reliability of the ML to a human agent capable of re-interpreting the outputs in light of background knowledge. Examined in isolation, ML systems whose working principles are solely based on JBDA cannot sustain claims about justification. 
What the examples of BenevolentAI and BenevolentAI* show, I believe, is that we might still be justified under JBDA-like algorithms if (a) the algorithm -- as a whole or as constituent parts -- implements scientific models, theories, principles, categories, and/or other elements purposed in our corpus of scientific knowledge, or (b) its outputs are later assessed by the relevant community and within a corpus of scientific knowledge (more on this in section \ref{RI3}.)%\footnote{Here, I follow the many that believe that scientific practice is crossed across by several (possibly non-scientific) abilities, such as inferring using analogical relations (e.g., between theories \cite[3127]{Seselja2014}), and the use of metaphors that inspire creative responses that cannot be rivaled by literal language but which represent a gap in our knowledge \cite[111]{BailerJones2009}.} 

Let me further illustrate this reliability indicator. Take again BenevolentAI, which utilizes information gathered from scientific research papers, structured and unstructured biomedical data, and drug and pharmaceutical industry data. BenevolentAI also implements knowledge graphs that structure data into causal relationships between known diseases, genes, environmental factors, and approved drugs \cite{Smith.Oechsle.ea2021}. Furthermore, BenevolentAI aligns with auxiliary assumptions, theories, and structures of drug molecular profiles, as well as mechanisms integral to the process of damaging healthy cells and tissues. It also incorporates theories about genetics, medical studies of disease, and biological models relating genes to drug effects. Through this knowledge-based integration, outputs produced by BenevolentAI are better justified than those by BenevolentAI*. Indeed, this JBDA-like version does not implement any accepted model or concept into its algorithm, does not operationalize knowledge graphs, and does not represent mechanisms integral to knowledge about damaging healthy cells and tissues. It is the theoretical rigor, along with a history of (un)successful implementations \cite{Duran.Formanek2018}, domain and expert competence, and possibly some skilled insight that leverages the reliability of BenevolentAI over and above BenevolentAI*.\footnote{Thanks go to Emanuele Ratti for pressing on clarifying this point.} %To put the same idea rather differently, the lack of such token$_2$ reliability indicators put BenevolentAI* is at a justificatory disadvantage compared to BenevolentAI.

%BenevolentAI*, nonetheless, classifies baricitinib* as having high probabilities of successfully combating COVID-19 symptoms. Why to believe in baricitinib*? Possibly because it can later be shown that this output is in agreement with background knowledge about inhibiting AAK1 and JAK 1/2 signaling pathways essential for combating the symptoms of COVID-19. 

%Integrating algorithms and their outputs with larger corpuses of knowledge means, inter alia, that principles of scientific methodology have been put to work. %, that outputs can be unified, explained, and understood as any other piece of scientific inquiry.

%because the algorithm implements these knowledge through multiple-level relationships between entities in the knowledge graph that reflect rich complexities of biomedical information, including causal relationships, pathways, processes, group memberships, ontologies, and hierarchies \cite{Smith2021}. 

\subsubsection{Type$_3$-RI: Social construction of reliability} \label{RI3}

The performance of algorithms (type$_1$-RI) is undoubtedly required for justification. But while a necessary condition, it is certainly not sufficient. The paradigmatic example is Wu and Zhang's CNN. This algorithm leverages a subset of RI$_1$ (most prominently, validation) but makes claims about an alleged law of facial recognition that is hard to accept. Expert and knowledge integration (type$_2$-RI) are also fundamental to the reliability of algorithms. But again, necessary but not sufficient. BenevolentAI furnishes a good example. The algorithm is robust and built on a solid scientific basis. However, baricitinib was later flagged as counter-prescribed for immunocompromised patients. So, what is missing? To my mind, the reliability of algorithms must also be assessed within social processes that aim to achieve standards of scientific value and thresholds for acceptance of \=o. Let me put the same idea in different form. The performance of algorithms along with expert knowledge and knowledge-based integration observe that the relevant scientific structures and processes, commitments and categories, entities and concepts are correctly implemented into, and computed by the algorithm. But justification of \=o requires something else. We need to further seek for coherence and consistency of \=o with a larger corpus of knowledge. In this way, we observe that the output is in agreement with accepted scientific commitments, standards of quality, evidence, and relevance, among other scientific qualifications. 

%Consider Wu and Zhang's CNN once more. One could make claims of justification that a photo is of a \o = \{criminal\} by either: (a) indicating that the algorithm implements specific structures and categories (e.g., definitions of criminality, instances of conviction, lawful classifications of crime), or (b) \=o is assessed by forensic experts, is consistent with our current theories of criminality (social, psychological, economic, legal), and the like. Now, neither strategy is applied by Wu and Zhang, and for this reason their CNN is a non-starter for scientific purposes. The algorithm does not implements any accepted theory or model of criminality --if anything, the algorithm implements a discredited theory like phrenology--, nor \=o is reviewed in light of scientific evidence about criminality. What about BenevolentAI? One could make claims of justification that \=o = \{barinitricib\} combats COVID-19 symthoms by either: (a) indicating that the algorithm implements the right causal relations, a wealth of medical knowledge, etc. or (b) \=o is assessed by forensic experts, is consistent with our current theories of criminality (social, psychological, economic, legal), and the like. 

Let me quickly illustrate how this reliability indicator would work using the story behind BenevolentAI. After announcing barinitricib, diverse groups within the scientific community began to debate the benefits --and dangers-- of this drug. A major concern that emerged was that the mechanisms of action of baricitinib would block JAK-STAT signaling pathways (mainly mediated by JAK1 and JAK2), thus impairing interferon-mediated antiviral responses \cite[1013]{Favalli.Biggioggero.ea2020}. Blocking interferon would allow attack by other viruses (e.g., herpes zoster and herpes simplex) which in some cases may be more harmful than COVID-19. Favalli and colleagues \cite{Favalli.Biggioggero.ea2020} later reported on the potential harms of administering baricitinib to some patients, most importantly immunodeficient patients. As a response, Richardson and colleagues accepted the conditions under which BenevolentAI was a reliable indicator --and reasonably use this debate and incorporate new functions into the algorithm.

As a reliability indicator, this scientific debate regulated on how much evidence was required to accept BenevolentAI's outputs. It draw thresholds of which errors and artifacts can be tolerated, and to what extent. It also determined which assumptions are fit for purpose. Commitments to reliable algorithms are commitments to a network of scientific methodologies, standards, and traditions expressed through scientific debate. As Catherine Elgin points out, this network enables scientists to build on each other's work. They can be confident that justified outputs have the epistemic value their discipline prescribes \cite[77]{Elgin1996}. Of course, disputes and disagreements among community members are to be expected. There may be conflicts over values, methods, and what constitutes acceptable evidence. Take again Favalli's concerns about administering baricitinib to a specific group of patients. Whereas Richardson largely agreed with Favalli's concerns, research on the drug continued. Further laboratory testing confirmed Richardson's beliefs.

The social formation and justification of beliefs is a complex enterprise that is not always successful. There are situated background assumptions and perspectives. Scientific inquiry is permeated by contextual values and interests. Many concepts built into algorithms are socially constructed, generational, and discipline-idiosyncratic. Take again variations in the definition of 'health' and 'disease' \cite{Boorse2011, Sisti.Caplan2017}. Each concept operates under a myriad of cultural, political, economic, and moral values. Caruana and colleagues \cite{Caruana.Lou.ea2015} discuss a neural network specified to predict pneumonia risk scores in patients and their readmission to hospital. Caruana et al. find that asthmatic patients are at low risk and thus less likely to require hospitalization than other patients (with chronic lung disease, for instance). Caruana et al.'s finding is statistically accurate (type$_1$-RI and type$_2$-RI indicators are present). However, the system is perceived as unreliable given that physicians have different starting assumptions about what a predictive algorithm should provide. According to Theunissen and Browning \cite{Theunissen.Browning2022}, physicians assume that the algorithm is predicting outcomes according to a shared baseline of care rather than differential care relative to the background of the patient. The ML outputs can be called into question because of oversimplifications in one or more assumptions in the system. This further shows that neither type$_1$-RI nor type$_2$-RI are individually—or jointly—sufficient for the reliability of algorithms.

% social construction of reliability is NOT about the RI, or the ML, or the assumptions of the system. Instead, it is the scientific debate about accepting computer-based results

CR makes an effort to foster belief-forming methods that accommodate social interventions, scientific scrutiny, and inter-domain justifications. Beliefs are justified in relation to a network of interconnected scientific beliefs. Yet, we cannot expect it to be recurrently successful. Contingent values and interests within the scientific community also find their way into the use and perception of the outputs of algorithms. Just like many other scientific methodologies, entrenching the reliability of algorithms requires a delicate balancing act. We bring together our best technical knowledge, theories, methodologies, and social skills. We do so in an attempt to justify believing that the output of a given ML system can be scientifically valuable, true, or can be epistemically trusted.

\section{Final thoughts} \label{final_remarks}

I have presented CR as a reliabilist epistemology for the justification of algorithms' outputs. I also presented and discussed diverse types and token reliability indicators that form the basis of reliable algorithms. In sum, I set out to defend the following claim: we have --or increasingly have-- justification for believing algorithms' outputs when they are rendered by a reliable belief-forming method. To be reliable here means that the algorithm is specified, coded, used, and maintained utilizing specific, tailor-made reliability indicators designed for the purpose at hand.

%I presented three families of reliability indicators capable of crediting such reliability to ML -- many of which are tailor-made to specifics of the ML.

%Reliability is achieved by off-loading judgment to rules, standards and techniques whose proper application ensures the acceptability of results (Elgin, TE, 166)
%``Even if the world's contribution is fixed, different accommodations to our inevitably limited resources may be equally reasonable'' (Elgin, TE, 88)

Admittedly, CR construes justification as inherently provisional. As discussed, reliability indicators might change over time and even be replaced. They might also be hard to measure or resolve conflicts. But I believe this is part of the self-critical and self-correcting endeavors that we find in scientific research. It might also be the best epistemic effort we can offer given a context and our limited resources. These activities constitute our best knowledge, metrics, methodologies, and practices, even if subjected to further scrutiny and revision. %In other words, justification does not derive from securing privileged information about the algorithm and its \textbf{internal} workings, but it is external, holistic, and perspectival.

I also listed a few issues with CR that I was unable to address but which might be considered conditional to its acceptability. I can accept that. But CR is a step in the right direction, if not as a more adequate epistemology of algorithms, at least as an alternative to internalist epistemologies. In this regard, the chapter has achieved its goal of setting up an externalist epistemology of algorithms, a goal that should be appreciated in its own right.

\section{Acknowledgments}

This paper has been in the making for quite some time. This means that many people need to be acknowledged and thanked for their diverse comments and suggestions. Let me begin with my co-editor. Thank you, Giorgia, for the endless close readings of this paper. Thanks also go to Jack Casey for many interesting discussions around these topics and helping me to avoid several mistakes. Many other people deserve recognition: Federica Russo, Emanuele Ratti, Edoardo Datteri, Giuseppe Primiero, Viola Schiaffonati, Rawad El Skaf, Manuel Barrantes, Karin Jongsma, Andrea Ferrario, Nico Formanek, Charles Rathkopf, and Atocha Aliseda. Some of them believe in CR, some of them don't. But they all have been supportive and encouraging of my ideas. Naturally, all wrongs—and rights—are mine. Finally, thanks go to Kass and Diego. Kass for being supportive beyond my comprehension. Thank you so much. And Diego, for teaching me an absolute truth: that everything is silly.

\end{document}